\documentclass[conference]{IEEEtran}

\usepackage{cite}
\usepackage{graphicx} 
\usepackage{algorithm}
\usepackage{algorithmic}
\usepackage[utf8]{inputenc}
\usepackage[T1]{fontenc}
\usepackage[english]{babel}
\usepackage{enumerate}
\usepackage{amsmath}
\usepackage{amssymb}
\usepackage{longtable}
\usepackage{pdfpages}
\usepackage{hyperref}
\usepackage{color}
\usepackage{algorithm}
\usepackage{algorithmic}

\def\v{\ensuremath{\mathbf{v}}}
\def\g{\ensuremath{\mathbf{g}}}
\def\x{\ensuremath{\mathbf{x}}}

\def\d{\ensuremath{\mathbf{d}}}
\def\p{\ensuremath{\mathbf{p}}}

\def\R{\mathbb{R}}

\def\A{\mathbf{A}}

\def\z{\mathbf{z}}

\def\u{\mathbf{u}}
\def\0{\mathbf{0}}

\DeclareMathOperator*{\argmin}{arg\,min}

\newcommand{\citet}[1]{{\cite{#1}}}

\newcommand{\GGCorr}[2]{{\color{black}#2}}

\newcommand{\GGCorrRev}[2]{{\color{black}#2}}
\newcommand{\GGmodRev}[1]{{\color{black}#1}}

\ifCLASSINFOpdf
            \else
                  \fi

\begin{document}
\title{Importance Sampling Strategy  for 
Non-Convex Randomized Block-Coordinate Descent }

\author{\IEEEauthorblockN{R\'emi Flamary}
\IEEEauthorblockA{Lagrange, UMR CNRS 7293, OCA\\Universit\'e C\^ote d'Azur\\
France\\
 remi.flamary@unice.fr}
\and
\IEEEauthorblockN{Alain Rakotomamonjy}
\IEEEauthorblockA{LITIS Rouen/LIF Marseille\\
Universit\'e de Rouen - Universit\'e Aix-Marseille\\
France \\
 {alain.rakoto@insa-rouen.fr}}
\and
\IEEEauthorblockN{Gilles Gasso\\ LITIS}
\IEEEauthorblockA{INSA de Rouen\\
France\\
 gilles.gasso@insa-rouen.fr}
}

\maketitle

\begin{abstract}
As the number of samples and dimensionality of optimization problems related
to \GGCorrRev{statistic}{statistics} an machine learning explode, block
coordinate descent algorithms have gained popularity \GGCorr{since they
are able to}{since they} reduce the original problem to several smaller ones.
Coordinates to be optimized are usually selected randomly according
to a given probability distribution. \GGCorr{In this work, we}{We} introduce an importance
sampling strategy that helps randomized coordinate descent algorithms
to focus on blocks that are still far from convergence. The framework 
applies to problems composed of the sum of two possibly non-convex
terms, one being separable and non-smooth. We have compared our algorithm to
a full gradient proximal approach as well as to a randomized block
coordinate algorithm that considers uniform sampling and cyclic block
coordinate descent. \GGCorr{Our
experimental results on toy and real-world problems,}{Experimental evidences} show the clear benefit of using an importance sampling strategy. 
\end{abstract}

\IEEEpeerreviewmaketitle

\section{Introduction}
In the era of Big Data, current computational methods for statistics
and machine learning are challenged by size of data  both
in \GGCorr{term}{terms} of dimensionality and \GGCorr{in terms of}{\!} number of examples.
Parameters of estimators learned from these large amount of data are usually
obtained as  minimizer of a regularized empirical risk problems of the form
\begin{equation}
  \label{eq:problem}
  \min_{\x \in \R^d} \{F(\x)=f(\x) + \lambda h(\x)\}
\end{equation}
where $f$ is usually a smooth and non-convex function
with Lipschitz gradient and
$h$ a non-smooth function.  In such a large-scale
and high-dimensionality  context,  most prevalent
approaches use first-order method based on gradient
descent  \cite{beck09:_fast_iterat_shrin_thres_algor} although second-order quasi-Newton algorithms have been considered \cite{rakotomamonjy2015dc}.

More efficient algorithms can be considered for solving problem
(\ref{eq:problem}) if $f$ and $h$ present some special structures.
When $h$ is separable, Problem \ref{eq:problem} can be expressed as
$$
 h(\x) = \sum_{i=1}^{m} h_i(\x_i) 
$$
\GGCorr{where $\x \in \R^d$. We suppose that $\x$}{We suppose that $\x \in \R^d$} is of the form $\x=[\x_1^\top,\dots\x_m^\top]^\top$  where $m$ is the number of groups in $\x$ and $\x_i \in \R^{d_i}$
and $\sum_i d_i = d$.
{In this case, methods that can use the group structure
such as }
coordinate descent algorithms \cite{Tseng2001} or randomized
coordinate descent \cite{nesterov2012efficiency} are {among} the
most efficient \GGCorr{methods}{ones} for solving
problem (\ref{eq:problem}).

In this paper, we focus on a specific class of randomized block proximal
gradient algorithm, useful when each block $h_i$ has a
special structure. We
suppose that each $h_i$ is a difference of convex functions and is non-smooth. However, it  has to have a closed-form
proximal operator \cite{gong2013jieping}. Such a situation mainly arises  when $h(\x)$ is a non-convex
sparsity-inducing regularizer. Common
non-convex and {non-differentiable} regularizers are the SCAD regularizer \cite{Fan_LI_scad_2001}, the $\ell_p$
regularizer \cite{KnightFuAsymptotics}, the capped-$\ell_1$  and the \textit{log} penalty \cite{CandesReweighted2008}. These regularizers have been 
frequently used for feature selection or for 
obtaining sparse models in machine learning
\cite{laporte13:_noncon_regul_featur_selec_rankin,gasso09:_recov_spars_signal_with_certain,CandesReweighted2008}.

A large majority of works dealing with randomized block coordinate descent algorithms (RBCD) considers uniform distribution of sampling \cite{nesterov2012efficiency,shalev2011stochastic,richtarik2012efficient}. 
Few attentions have been devoted to the use of arbitrary distribution \cite{lu15:_random_nonmon_block_proxim_gradien_method,qu2014coordinate}. In these two latter efforts, principal statement  is that the probability of drawing any block should be not less than a $p_{min}>0$ value {to ensure that
all blocks have non-zero probabilities to be selected} and hence to guarantee convergence in expectation
of the algorithm. 
 However, because no prior knowledge are usually available for directing the choice of the probability distribution of block sampling, experimental analysis of the randomized
algorithms usually consider uniform distribution.

\GGCorr{Our contribution in this work is to propose}{This paper proposes}  a probability distribution
for randomized block coordinate sampling that goes beyond the uniform
sampling and that is updated after each iteration of the \GGCorr{RBCD}{\!} algorithm. \GGCorr{Hence}{Indeed}, we have designed a distribution that  is dependent on approximate optimality condition of the problem. Owing to such a distribution, described in Section \ref{sec:framework} we can bias the sampling
towards coordinates that are still far from optimality allowing to save 
\GGCorrRev{subtantial}{substantial} computational efforts as illustrated by our empirical
experiments (see Section \ref{sec:expe}).

\section{Framework and algorithm}
\label{sec:framework}

\subsection{Randomized BCD}

We discuss now a generic approach for \GGCorr{ the optimization of problem
\eqref{eq:problem}. We focus, in this case, on problems where}{solving problem
\eqref{eq:problem} when}  
$h(\cdot)$ is separable \GGCorr{and we want to take}{by taking} advantage of
 this separability. \GGCorr{First, we define $\nabla_if(\x)$ as the partial gradient
{at $\x$} of
$f(\cdot)$ with respect to $\x_i$.}{The general framework is shown in Algorithm \ref{alkgo:bcd} where $\nabla_if(\x)$ is the partial gradient
{at $\x$} of $f$ with respect to $\x_i$.}

\begin{algorithm}[t]
  \begin{algorithmic}[1]
     \STATE Set initial $\x^0$, $\theta>0$, $\eta>1$, $\sigma >0$
      \FOR{$k=1,2,\dots$} 
      \STATE{$i\leftarrow $ randomly select current block from $\{1,2,\dots,\GGCorr{p}{m}\}$} according to a probability distribution $\p$ 
      \STATE{$\d\leftarrow \mathbf{0}; \d_i\leftarrow\nabla_if(\x)$}
      \STATE{ $\x^k\leftarrow
        \text{prox}_{\frac{1}{\theta_k}h}(\x^{k-1}-\frac{1}{\theta_k}\d)$,$j\leftarrow
        0$}
      \WHILE{$F(\x^k)>F(\x^{k-1}) \GGCorrRev{+}{-} \frac{\sigma}{2}\|\x^k-\x^{k-1}\|$}

      \STATE{$j\leftarrow j+1$} \GGmodRev{and set $\gamma = (\eta)^j$}
      \STATE{$\x^k\leftarrow \text{prox}_{\frac{1}{\GGCorrRev{\theta_k  \eta^j}{\theta^k \gamma}}h}(\x^{k-1}-\frac{1}{ \GGCorrRev{\theta_k  \eta^j}{\theta^k \gamma} }\d)$} 
      \ENDWHILE
      \ENDFOR
  \end{algorithmic}
  \caption{Randomized Block Coordinate Descent (RBCD)}
  \label{alkgo:bcd}
\end{algorithm}

 At each iteration in the algorithm a block $i$ is selected to be
optimized (line 3).  Then, a partial proximal gradient
step is performed (line 5) for the selected group. It consists in
solving efficiently the {proximal operator}
$$\text{prox}_{\frac{1}{\theta}h}(\v)=\argmin_\x
\frac{1}{2}\|\x-\v\|^2+\frac{1}{\theta} h(\x) .$$
Note that since $h$ is separable, the proximal operator can be applied only on
the current group $i$ and will update only $\x_i$. 
 A backtracking (line 6-9) may be necessary to ensure a
decrease in the objective $F$ but a non monotone version can also be
used as discussed in
\cite{lu15:_random_nonmon_block_proxim_gradien_method}. Finally, if
the number of groups is set to 1, then the algorithm boils down to
GIST \cite{gong2013jieping}, \emph{i.e.} a proximal method for non-convex optimization.

This randomized algorithm is interesting \emph{w.r.t.} the classical
proximal gradient descent since  it does not require the
computation of the full gradient at each iteration. For instance, when estimating a
linear model, the loss $f$ can be expressed as $f(\x)=L(\A\x)$. The
gradient is \GGCorr{of the form}{\!} $\nabla f(\x)=\A^\top L'(\A\x)$ where the derivative
$L'$ is computed pointwise. \GGCorrRev{When computing}{Computing} the partial gradient  $\nabla_i f(\x)=\A_i^\top L'(\A\x)$ where
$\A_i$ is \GGCorr{the matrix  containing only the columns}{the submatrix} of $\A$ corresponding to
group $i$  \GGCorrRev{, which}{}
requires much less \GGCorr{operations. Indeed,  Table \ref{tab:tabcomplexity}
reports the floating operations required at each iteration for
both algorithms. Note that the
complexity per iteration is much smaller for RBCD since $d_i\ll
d$.}{floating operations as reported in Table \ref{tab:tabcomplexity} since $d_i\ll
d$.} In addition, this computational complexity can be greatly
  decreased by storing \GGCorr{at each iteration the prediction $\A\x$ and  by
using the low complexity update $\A_i(\x^k_i-\x^{k-1}_i)$ at each iteration. 
}{the prediction $\A\x$ and  by
using the low complexity update $\A_i(\x^k_i-\x^{k-1}_i)$ at each iteration. 
}

\begin{table}[t]

  \caption{Floating operation at each iteration for the GIST and RBCD
    for a linear model of the form $f(\x)=L(\A\x)$. $d_i$ is the
     dimensionality of the group $i$ updated at the current iteration.}
  \centering
  \label{tab:tabcomplexity}
  \begin{tabular}{|l|c|c|}\hline
    Task & GIST & RBCD\\\hline
    Gradient computation & $2nd+n$ & $2nd_i+n$\\
    Proximal operator & $d$ & $d_i$\\
    Cost computation & $nd+n$ & $nd_i+n $\\\hline
  \end{tabular}

\end{table}

\subsection{Block selection and importance sampling}

The convergence of the RBCD algorithm is clearly dependent of the
block selection strategy of line 3 in Algorithm \ref{alkgo:bcd}. One
can select the group using classic cyclic rule as in
\cite{chouzenoux13:_variab_metric_forwar_backw},\GGmodRev{\cite{bolte2013proximal}} or using the realization
of a random distribution  \cite{nesterov2012efficiency},\GGmodRev{\cite{2013Tappenden}}. The uniform
distribution is often used in order to
ensure that all blocks are updated equally, but convergence in
expected value has been proved for any discrete distribution that have
non-null components, ($p_{min}>0$) \cite{lu15:_random_nonmon_block_proxim_gradien_method}. 

In this work, we introduce a  novel probability distribution for
sampling blocks in RBCD. This distribution is dependent
on the optimality conditions of  each block. 
In other words, we want to update more often 
blocks that are still far from convergence. Formally, let $\p\in\R^{+m}$ be the discrete density distribution such that $p_i$  is the probability that the
block $i$ is selected at a given iteration and $\sum_i p_i=1$. 
We propose in this work to use the following distribution
\begin{equation}
p_i=\frac{\epsilon+(1-\epsilon)\frac{z_i}{\|\z\|_\infty}}{m\epsilon+(1-\epsilon)\frac{1}{\|\z\|_\infty}\sum_i z_i}\label{eq:p}
\end{equation}
where \GGCorr{$\epsilon>0$}{$\epsilon \in (0, \, 1 ]$} is a user-defined parameter, $\z\geq \mathbf{0}$ is a vector composed of coordinates $\{z_i\}_{i=1}^m$ and $\|\z\|_\infty=\max_i|z_i|$ is the infinite norm. 
As made clearer in the sequel, a component $z_i$
encodes the
optimality condition violation in each block. Indeed,
let $h_i= h_{i,1} - h_{i,2}$, with $h_{i,1}$ and $h_{i,2}$ being
two convex functions, then if $\x^\star$ is a local
minimizer of $F(\x)$,  from
Clarke subdifferential calculus \cite{rockafellar2009variational}, one
can show that a necessary condition of optimality is
that there exists $\v\in \partial h_{i,1}(\x^\star)$ and 
$\u\in \partial h_{i,2}(\x^\star)$ such that
 $0 \in \nabla_i f(\x^\star)+\lambda\v - \lambda \u$ for all $i$.
 Accordingly, we define the optimality condition violation 
$z_i$   as
\begin{equation}
z_i=\min_{\v\in \partial h_{i,1}(\x), \u\in \partial h_{i,2}(\x)}\|\nabla_i f(\x)+\lambda\v - \lambda \u\|_\infty\label{eq:kktviol}
\end{equation}
The role of $\epsilon$  in Equation (\ref{eq:p}) is to balance the
effect of the optimality condition on the distribution. When
$\epsilon=1$, we retrieve a uniform distribution. Other values of
$\epsilon$ will ensure that if a variable in a block has not
converged, its block is likely to be updated more often than a block
that has converged. Note that, owing to the DC decomposition
of $h$, the  violation \eqref{eq:kktviol} can be
easily computed, even for
non-convex penalty function such as SCAD or the log-sum as discussed in
\cite{boisbunon2014active}.

Computing the optimality condition violation vector 
$\z$ is not
possible in practice {for RBCD
since it requires the full gradient of the problem, which
as discussed in the previous section, is not computed at each
iteration}. 
\GGCorrRev{}{As a solution, we propose to use a vector $\tilde \z$ initialized with the exact condition violation computed from the initial vector $\x^0$. Thereon,  only the $i^{\mbox{th}}$ entry of $\tilde \z$ is updated at each iteration leading to an approximate optimality condition evaluation. Indeed, in algorithm
\ref{alkgo:bcd} line 4, {when a partial gradient $\nabla_i f(\x)$ is computed, we can use it to update the approximate $\tilde z_i$ and then update the probabilities $\p$ accordingly.}}
This latter vector is clearly a
coarse approximation of the optimality condition violation but as shown in the
experiments \GGCorr{ai}{it is} a relevant choice for the \GGCorr{importance sampling scheme
described above.}{proposed importance sampling scheme.}

\subsection{On tricks of the trade}

The proposed optimization algorithm has an important parameter that has
to be chosen carefully: the initial gradient \GGCorrRev{step $\theta^k$}{step size $1/\theta^k$} at each
iteration. If chosen  too small, the gradients steps will barely
improve the objective value, if chosen too large the backtracking step
in lines 6-9 will require numerous computation of the loss
function. In this work we use an extension of the Barzilai-Borwein
(BB) rule  that has been proposed in a non-convex scheme by
\cite{gong2013jieping}. \GGCorr{This approach consists in estimating the
Hessian matrix of the problem by a weighted identity matrix, and using
a Newton step with this approximate matrix.}{This approach consists in  using
a Newton step with the approximate Hessian $\sigma \mathbf{I}$.} When performing the full
gradient descent in GIST, the BB rule gives 
\begin{equation}
\theta^{k+1}=\frac{\Delta \x^\top \Delta \g}{\Delta \x^\top \Delta
  \x}\label{eq:bbrule}
\end{equation}
where $\Delta \x=\x^k-\x^{k-1}$ and $\Delta \g=\nabla f(\x^k)-\nabla
f(\x^{k-1})$. Again, in our algorithm the full gradient is not
available but we can still benefit from the second-order
approximation brought to us by the BB rule. We propose to this end to
model the Hessian as a diagonal matrix where the weight of the
diagonal is block-dependent. In other word, we store an estimate
$\boldsymbol{\theta}\in\R^{+m}$ whose components  $\theta_i$ are updated
similarly to equation \eqref{eq:bbrule} but using instead partial gradient and variations $\Delta \x_i=\x^k_i-\x^{k-1}_i$ and $\Delta \g_i=\nabla_i
f(\x^k)-\nabla_i f(\x^{k-1})$. This new rule is actually more general
than the classical BB-rule since  it brings
local information and  encodes a more precise Hessian
approximation with group-wise coefficients similar to the variable
metric in \cite{chouzenoux13:_variab_metric_forwar_backw} .

\section{Numerical Experiments}
\label{sec:expe}

\begin{figure*}[t]
  \centering
~\hfill
\includegraphics[width=.3\linewidth]{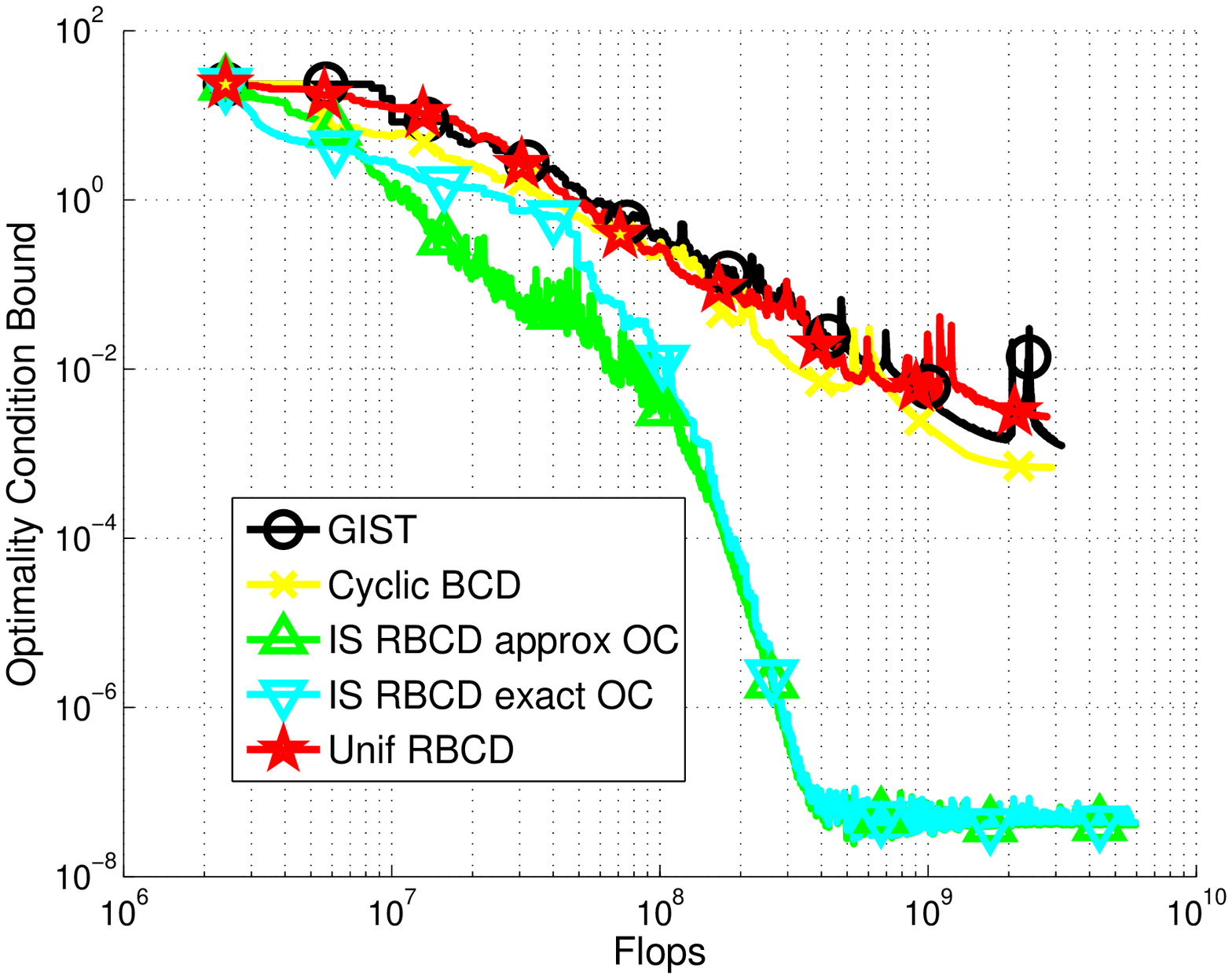}\hfill  
\includegraphics[width=.3\linewidth]{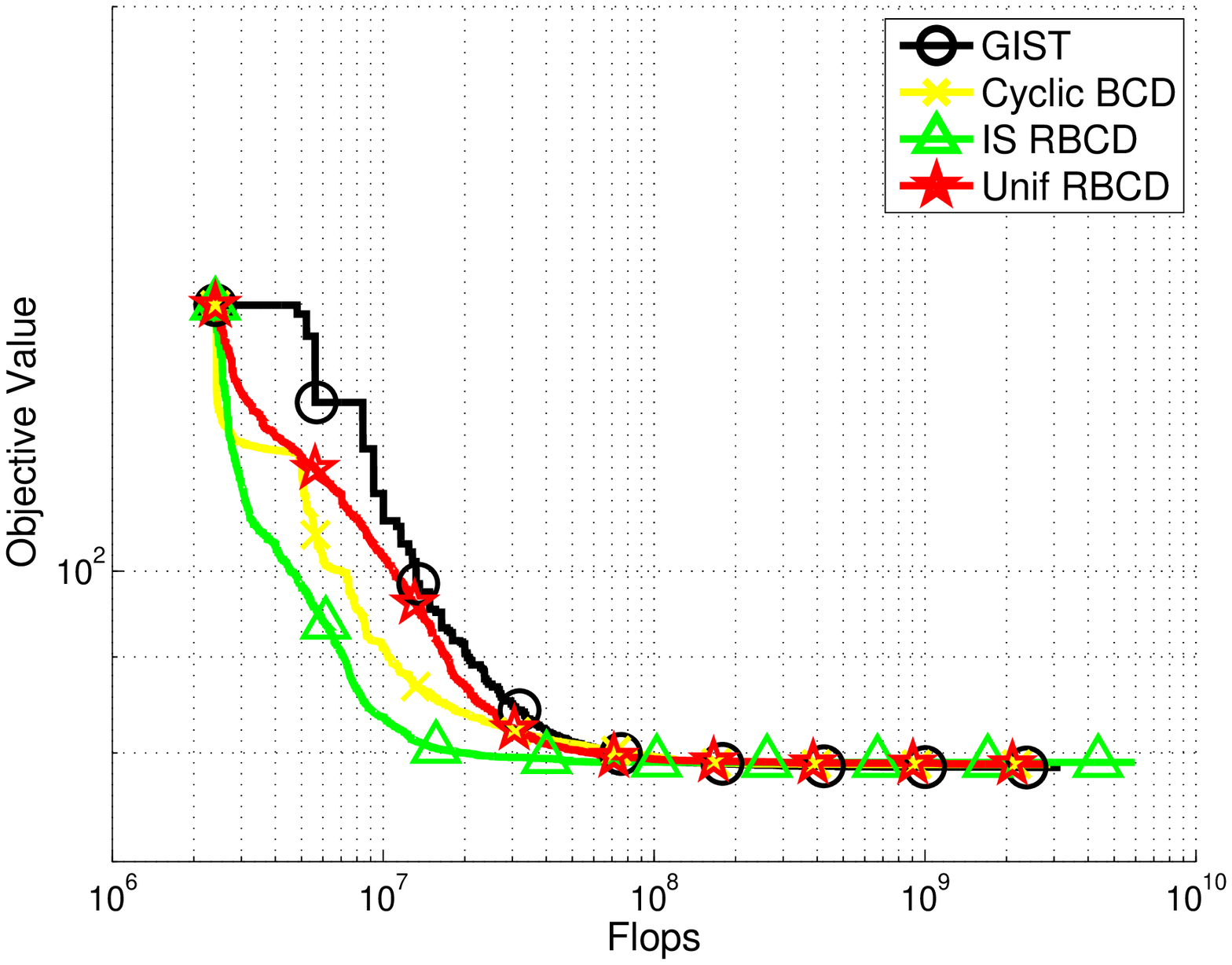}\hfill~
\includegraphics[width=.3\linewidth]{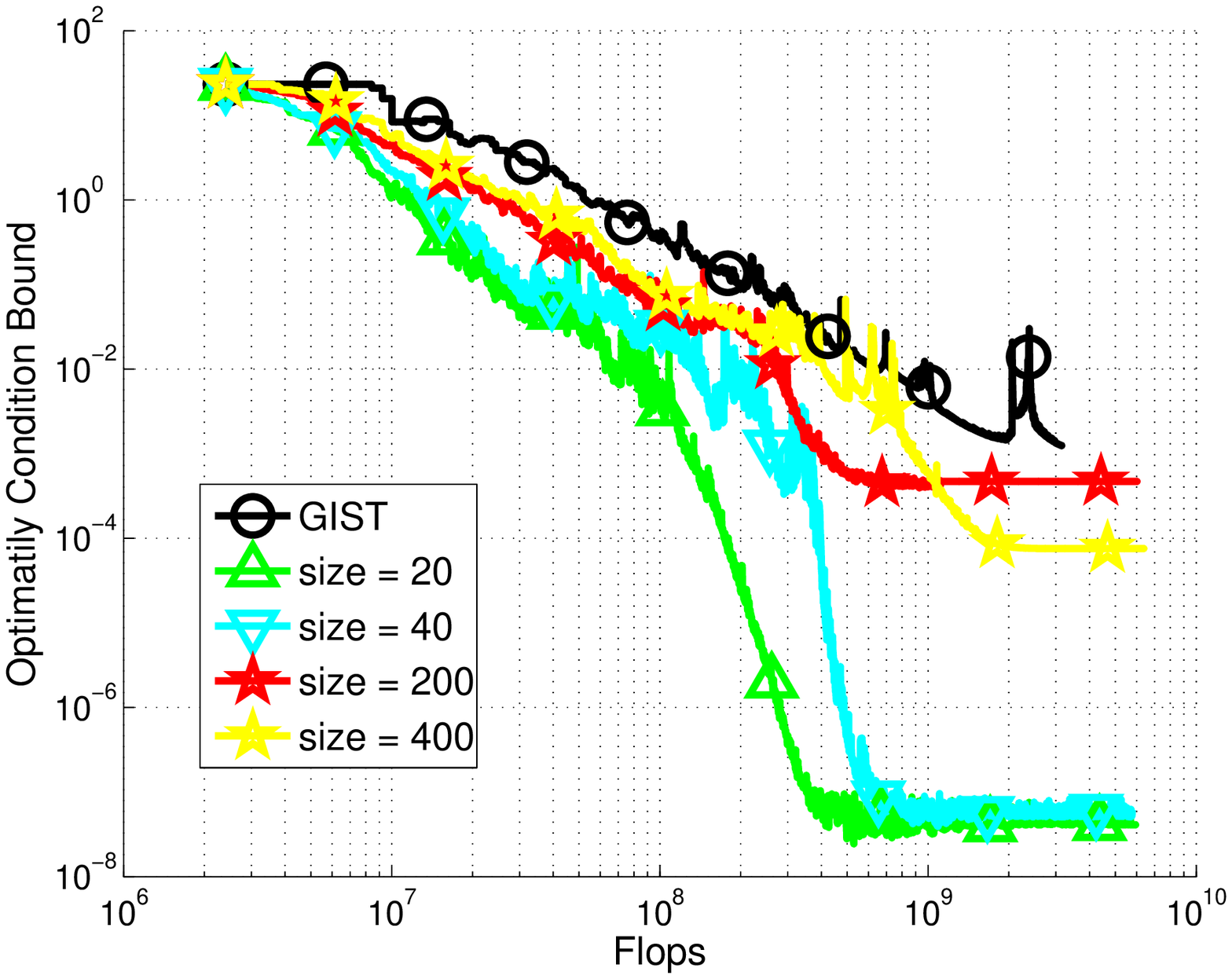}\hfill~
\caption{Example of (left) optimality condition violation and (middle)
  objective value evolution with respects to the number of flops,  averaged
over $20$ iterations and with blocks of size $20$. For the left panel,
we have plotted the exact violation (computed with $\z$) as well as
the approximated one (computed with $\tilde \z$).
 (right) Optimality
conditions violation averaged
over $20$ iterations for different block sizes used in IS RBCD. (Best viewed in color)}
  \label{fig:examples}
\end{figure*}

In this section, we illustrate the behaviour of our 
randomized BCD algorithm with importance sampling on some toy
and real-world classification problems. For all problems, 
we have considered a logistic  loss function 
and the log-sum non-convex sparsity inducing penalty
defined as
$$
h(\x)=  \rho \sum_{i}^d \log \left(1 + \frac{|x_i|}{\rho} \right) 
$$
with $\rho>0$.
We have compared our algorithm to a non-convex proximal gradient
algorithm known as GIST \cite{gong2013jieping} and
a randomized BCD version of GIST with uniform sampling
\cite{lu15:_random_nonmon_block_proxim_gradien_method}. {Note
  that since this regularization term is fully separable per variable,
  we used a separation of $m$ blocks of size $\frac{d}{m}$ variables. }

\subsection{Toy problem}

\GGCorr{The toy problem is the same as the one used by \cite{rakotomamonjy2015dc}.
The task is a binary classification problem in $\R^d$. }{As in \cite{rakotomamonjy2015dc} we consider a binary classification problem in $\R^d$.}
Among
these $d$ variables, only $T$ of them define a subspace of $\R^d$
in which classes can be discriminated. For these $T$ relevant
variables, the two classes follow a Gaussian pdf with means
respectively $\boldsymbol{\mu}$ and $\boldsymbol{-\mu}$  and covariance matrices randomly
drawn from a Wishart distribution \GGmodRev{$W(\mathbf{I}, T)$ where $\mathbf{I}$ is the identity matrix}. \GGmodRev{The components of} $\boldsymbol{\mu}$ \GGCorrRev{has been randomly}{have been independently and identically} drawn
from $\{-1,+1\}$. The other $d-T$ non-relevant variables follow
an i.i.d Gaussian probability distribution with zero mean and
unit variance for both classes.  We have respectively sampled
$n$ and $n_t$ = 1000 number of examples for training and
testing. Before learning, the training set has been normalized
to zero mean and unit variance and test set has been rescaled
accordingly. Note that the \GGCorrRev{hyperparameters $\lambda$ and $\rho$  of}{hyperparameter $\lambda$ or any other parameters related to} the
regularization term  have been set so as to maximize the
performance of the GIST algorithm on the test set. We have
 initialized all algorithms with the zero vector ($\x^0 = \mathbf{0}$).

The different algorithms have been compared based on their
computational demands and more exactly based on the
number of flops they need for reaching a stopping
criterion. Hence, this criterion is  critical for a fair
comparison. The GIST algorithm has been run until it
reaches a necessary optimality condition $\|\z\|_\infty$ lower than $10^{-3}$ 
or until $1000$ iterations is attained.
For the randomized algorithms, including our approach
\GGCorr{with importance sampling}{\!}, the stopping criterion is set
according to a maximal number of iterations. This number
is set so that the number of coordinate gradient evaluations \GGCorrRev{are}{is} equal
for all algorithms i.e we have used the number of GIST iterations$\times
{m}
$ where $m$ is the number of {blocks}. In the sequel, the number
of flops reported is related to those needed for computing both
function values and gradient evaluations.

Figure \ref{fig:examples} (left)  presents some examples
of  optimality condition $\|\z\|_\infty$ evolution with respects to the number
of flops. These curves are obtained as averages over $20$ iterations
of the results obtained for a given experimental set-up (here $n=200$,
$d=2000$ and $T=20$). We can first note that with respect to optimality
condition,  RBCD algorithm with uniform sampling (Unif RBCD) behaves
similarly to the GIST algorithm and a cyclic BCD (\GGCorrRev{BCD Cyclic}{Cyclic BCD}). In terms of flops, few gain can be
expected from such an approach. 
Instead, using importance sampling (IS RBCD) considerably helps in
improving convergence.
Such a behaviour can also be noted when monitoring evolution of the objective
value (see central panel in Figure \ref{fig:examples}). Randomized algorithms tend to converge faster towards their optimal
value with a clear  advantage to the importance sampling approach. 
{Finally, while they are \GGCorr{note}{not} reported due to lack of space, the final
  classification performances  are similar
  for all three methods.}

Figure \ref{fig:examples} (right) depicts evolutions of optimality conditions
depending on block-coordinate group size. We can note that regardless
of this size, our importance sampling approach achieves better performance
than the GIST algorithm. In addition, it is clear that for our examples,
the smaller the size is, the faster convergence we obtain.

\subsection{Real-world classification problems}

\begin{table*}[t]
  \centering
  \caption{Comparison of GIST and randomized BCD algorithms on real-world benchmark problems. The first columns of the table provide the name of
the \GGCorr{datasets as well as }{datasets,} the number of training examples $n$
and their dimensionality $d$. Three measures of performances are provided :
the classification rate, the number of flops needed for convergence, the optimality condition. The objective value is given for a sake of information but it
is not a relevant criterion in a non-convex problem. }

  \begin{tabular}[h]{lcc|r|cccc}\hline
data & $n$ & $d$& Algorithm & Class. Rate (\%) & Flops $\times 10^9$ & Opt. Condition & Obj. Val \\
\hline
classic & 7094 & 41681 &GIST & 96.37$\pm$0.5 & 9277.76$\pm$64.6 & 0.03$\pm$0.0  & 32.64$\pm$2.2 \\
classic & 7094 & 41681 &IS RBCD & 95.11$\pm$0.7 & 347.16$\pm$4.1 & 0.01$\pm$0.0 & 25.23$\pm$0.8 \\
classic & 7094 & 41681 & Unif RBCD & 95.87$\pm$0.6 & 364.12$\pm$66.2 & 0.03$\pm$0.0 & 35.26$\pm$0.8 \\\hline
la2 & 3075 & 31472 &GIST & 91.11$\pm$1.1 & 3148.75$\pm$287.8 & 0.06$\pm$0.1  & 39.42$\pm$57.7 \\
la2 & 3075 & 31472 &IS RBCD & 90.98$\pm$1.2 & 101.16$\pm$3.6 & 0.15$\pm$0.2 & 43.35$\pm$59.0 \\
la2 & 3075 & 31472 & Unif RBCD & 91.04$\pm$0.9 & 108.11$\pm$4.8 & 0.23$\pm$0.3 & 45.51$\pm$59.0 \\\hline
ohscal & 11162 & 11465 &GIST & 88.30$\pm$0.6 & 7452.22$\pm$895.6 & 2.65$\pm$2.3  & 520.41$\pm$451.2 \\
ohscal & 11162 & 11465 &IS RBCD & 87.88$\pm$0.8 & 164.42$\pm$21.5 & 0.87$\pm$0.6 & 480.53$\pm$428.5 \\
ohscal & 11162 & 11465 & Unif RBCD & 87.75$\pm$0.8 & 156.45$\pm$17.7 & 1.14$\pm$1.1 & 480.55$\pm$428.5 \\\hline
sports & 8580 & 14870 &GIST & 97.93$\pm$0.4 & 5034.75$\pm$1219.5 & 0.11$\pm$0.1  & 208.11$\pm$215.2 \\
sports & 8580 & 14870 &IS RBCD & 97.76$\pm$0.5 & 154.74$\pm$20.3 & 0.07$\pm$0.1 & 212.05$\pm$215.3 \\
sports & 8580 & 14870 & Unif RBCD & 97.86$\pm$0.4 & 173.99$\pm$10.6 & 0.39$\pm$0.3 & 222.38$\pm$215.3 \\\hline

  \end{tabular}
  \label{tab:result_classif}
\end{table*}

We have also compared these \GGCorr{three}{\!} algorithms on real-world high-dimensional
learning problems. The related datasets have been already used as benchmark
datasets in \cite{gong2013jieping,rakotomamonjy2015dc}.
For these problems, we have used $80\%$ of the examples as training
set and the \GGCorr{rest }{remaining} as test set. Again, hyperparameters of the model
have been chosen so as to roughly maximize performances of the GIST algorithm. 
Stopping criteria of all algorithms have been set as \GGCorr{described above for the toy problems}{previously}. However, maximal number of iterations has been set 
to $5000$ for GIST. In addition, \GGCorr{here}{\!} we have limited the maximal number of iterations
to $20000$. {The number of blocks has been set to $m=100$ for all datasets.}

Performances of the different algorithms are reported in Table \ref{tab:result_classif}. Three measure of performances \GGCorr{can be}{have been} compared. Classification rates
of all algorithms are almost \GGCorrRev{on par}{similar} although \GGCorr{difference}{differences} in performances
are statistically significant in favor of GIST according to a Wilcoxon \GGCorr{signrank}{sign rank}
test with a p-value of $0.05$. We explain {this \GGCorr{slight advantage in favor of
GIST}{\!} by the fact that  regularization parameters have been
selected \emph{w.r.t.} to its \GGmodRev{generalization} performances}. The number of flops needed for convergence are
highly in favor of the randomized algorithms. The factor gain in flops
ranges in between $26$ to $45$. Interestingly, exact optimality conditions
after algorithms have halted are always in favor of our importance sampling
randomized BCD algorithms except for the \emph{la2} dataset. Note that,
in the table, we have also provided  the objective values of the \GGCorr{three}{\!}
algorithms
upon convergence. As one may have expected in a non-convex optimization problem,
 different ``nearly'' optimal objective values leads
to similar classification rate performances stressing the existence of
several local minimizers with good generalization property.

\section{Conclusion}
This paper introduced a framework for randomized block coordinate
descent algorithm that leverages on importance sampling. We presented
a sampling distribution that biases the algorithm to focus
on block coordinates that are still far from convergence. While this
idea is rather simple, our experimental results have shown that
it considerably helps in achieving a faster empirical convergence
of the randomized BCD algorithm.
Future works will be devoted to the theoretical analysis
of the importance sampling impact on the convergence rate. 
In addition, we plan to carry out \GGCorrRev{a thoroughful}{thorough} experimental
analyses that unveil the impact of the algorithm parameters.

\bibliographystyle{IEEEtranS}

\end{document}